\documentclass[review, 12p]{elsarticle}
\usepackage{amssymb}

\usepackage{caption}
\usepackage{multirow}
\usepackage[table,xcdraw]{xcolor}
\usepackage{hyperref}

\usepackage{natbib}
\usepackage{ragged2e}
\usepackage[final]{changes}
\hypersetup{
	colorlinks=True,
	citecolor=cyan,
	linkcolor=cyan,
	urlcolor=cyan
}

\begin{document}
	\begin{frontmatter}
		
		\title{Joint graph entropy knowledge distillation for point cloud classification and robustness against corruptions}
	
		\author[label1]{Zhiqiang Tian}
		\author[label1]{Weigang Li\corref{cor1}}
		\cortext[cor1]{Corresponding author}
		\ead{liweigang.luck@foxmail.com}
		\author[label1]{Junwei Hu}
		\author[label2]{Chunhua Deng}
		
		\address[label1]{School of Information Science and Engineering, Wuhan University of Science and Technology, Wuhan 430081, China}
		\address[label2]{School of Computer Science and Technology, Wuhan University of Science and Technology, Wuhan, 430065 , China}
		
		\begin{abstract}
			Classification tasks in 3D point clouds often assume that class events \replaced{are }{follow }independent and identically distributed (IID), although this assumption destroys the correlation between classes. This \replaced{study }{paper }proposes a classification strategy,  \textbf{J}oint \textbf{G}raph \textbf{E}ntropy \textbf{K}nowledge \textbf{D}istillation (JGEKD), suitable for non-independent and identically distributed 3D point cloud data, \replaced{which }{the strategy } achieves knowledge transfer of class correlations through knowledge distillation by constructing a loss function based on joint graph entropy. First\deleted{ly}, we employ joint graphs to capture add{the }hidden relationships between classes\replaced{ and}{,} implement knowledge distillation to train our model by calculating the entropy of add{add }graph.\replaced{ Subsequently}{ Then}, to handle 3D point clouds \deleted{that is }invariant to spatial transformations, we construct \replaced{S}{s}iamese structures and develop two frameworks, self-knowledge distillation and teacher-knowledge distillation, to facilitate information transfer between different transformation forms of the same data. \replaced{In addition}{ Additionally}, we use the above framework to achieve knowledge transfer between point clouds and their corrupted forms, and increase the robustness against corruption of model. Extensive experiments on ScanObject, ModelNet40, ScanntV2\_cls and ModelNet-C demonstrate that the proposed strategy can achieve competitive results. 
		\end{abstract}

	\end{frontmatter}

	\section{Introduction}
	
	\replaced{Owing }{Due }to the intuitive, adaptable, and memory\replaced{-}{ }efficient properties of point cloud\added{s}, its concept has a wide range of practical applications in computer vision, autonomous driving, robotics, and other related fields \hypersetup{citecolor=cyan}\cite{hu2022daniel, ma2022effective, zhang20233d}. Following the ground\added{-}breaking work of PointNet \cite{Pointnet}, a series of deep learning-based models \replaced{consider }{take }the performance of point cloud tasks to a higher level. However, the scale of point cloud data is much smaller than that of image dataset, and the annotation cost is expensive. Using existing data to train the model \deleted{will} lead\added{s} to overfitting and \added{a }lack of generalization ability \cite{PointNeXt}. In recent years, researchers have explored several data augmentation \cite{Pointmixup, Pointcutmix, RSMix} \replaced{strategies }{strategy }and training frameworks \cite{SE-SSD, CLpoint, PSD}\replaced{ that}{, these methods} assume \deleted{that }the class events are \deleted{drawn }independently and identically (IID)\added{ drawn}, \replaced{however }{but }this assumption is violated in several real\added{-}life problems, thus making it difficult for the model to learn the hidden relationship between classes in the training process.
	
	Although the\added{ IID} assumption \deleted{of IID }can simplify\added{ the} calculation\added{s} and reduce complexity, it is too idealistic and \deleted{will }ignore\added{s} the correlation between\added{ the} data. Dundar et al. \cite{dundar2007learning} pointed out that the assumption that data obey IID in classification tasks is contrary to \replaced{a }{the }real problem. Cao et al. \cite{cao2022beyond} demonstrate\replaced{d}{s} the limitations of \added{the }IID\deleted{'s} hypothesis, which has limitations and differences in knowledge and ability from the \replaced{actureal}{real }situation of \replaced{a }{the }real problem. In this \replaced{study}{paper}, we re-examine the point cloud classification task from the perspective of non-IID.
	\begin{figure*}[ht]
		\centering
		\includegraphics[width=\linewidth]{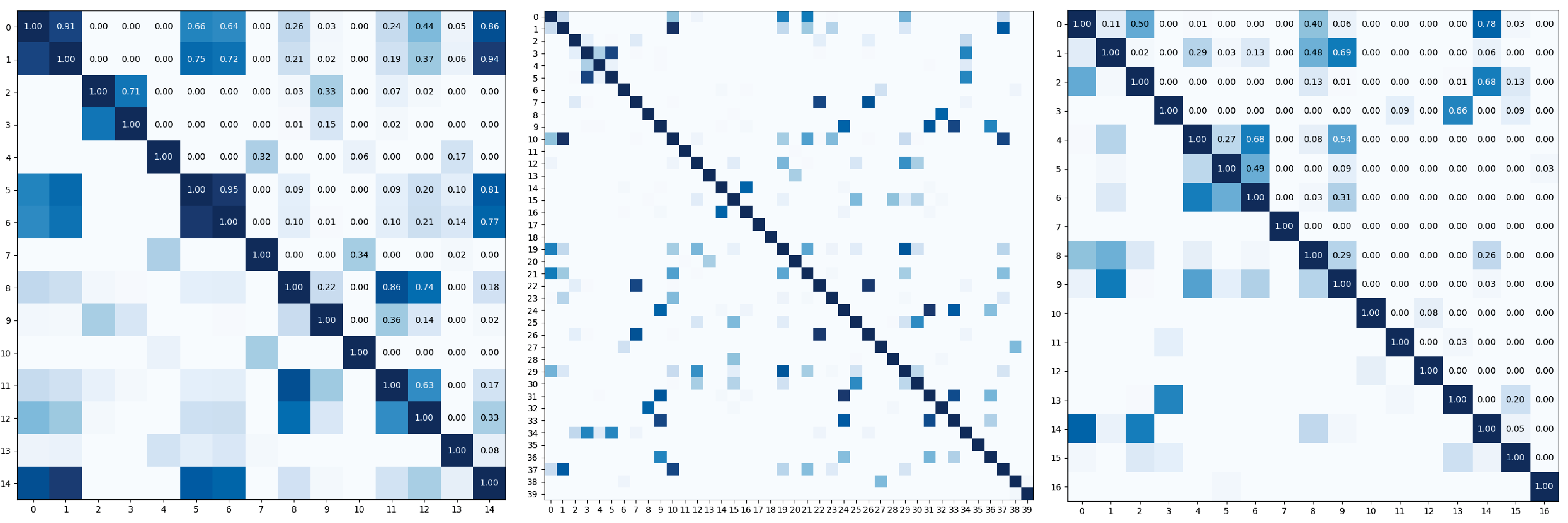}
		\caption{Class\deleted{es} correlation matrix on ScanObject, ModelNet40 and ScanNetV2\_cls (left to right) dataset\added{s}: \replaced{W}{w}e abstracted the features of each sample to obtain a tensor of 1 $\times$ 2048, and then randomly selected 20 samples for each class to obtain a matrix of $N \times$ (20 $\times$ 2048), where "$N$" denotes \added{the }number of classes. \added{A }T-test was employed to detect the correlation \replaced{between }{of }each class. The value in the matrix represent\replaced{s}{ed} the correlation score between each class.}
		\label{ttest}
	\end{figure*}

	In contrast to 2D images, 3D point clouds exhibit unordered and shape-invariant properties, which can \replaced{result in}{lead to }samples from different classes presenting similar representations after \deleted{undergoing }transformat\replaced{ing}{ions}.
	Fig. \hypersetup{linkcolor=cyan}\ref{ttest} shows the correlation matrices among classes in the point cloud datasets. This phenomenon suggests that events belonging to different classes follow non-IID, and \deleted{that there exist }implicit relationships\added{ exist} between classes.
	
	\replaced{We}{To accurately describe the latent relationships between classes, we} propose the Joint Graph Entropy Knowledge Distillation (JGEKD) strategy\added{ to describe the latent relationships between classes accurately}. First\deleted{ly}, a joint graph was constructed to describe the latent relationships between classes, and a graph entropy-based approach was proposed to characterize the information content of \added{the }correlations between non-\replaced{IID}{iid} classes. Subsequently, \deleted{we employed }knowledge distillation\added{ was employed} to train the model to achieve domain transfer between classes. However, \deleted{only }constructing inter\deleted{-}class correlations \added{alone }cannot make the model adaptable to diverse point cloud samples. To handle point cloud data \deleted{that is }invariant to spatial transformations, we construct \replaced{S}{s}iamese structures and develop two frameworks, joint graph entropy self knowledge distillation (JGEsKD) and joint graph entropy teacher knowledge distillation (JGEtKD), which are instrumental in learning the contextual information of the same sample in alternative representations. Finally, to address the common occurrence of corrupted point clouds in real-world scenarios, we propose an adversarial training strategy based on the aforementioned framework. A Siamese network \replaced{was }{is }constructed to simultaneously abstract standard and corrupted point cloud data, and JGEKD \replaced{was }{is }utilised to achieve domain transfer between the two.
	
	The primary contributions of this paper are outlined as follows:
	
	(1) A \added{JGEKD }strategy \replaced{was }{of is }proposed, which is characterized by the description of the relationship between classes through \added{a }joint graph, and the realisation of the model's learning of non-IID point cloud data through graph distillation.
	
	(2) Two straightforward yet effective frameworks, \deleted{namely }JGEtKD and JGEsKD, \replaced{were }{are }devised based on the JGEKD strategy to handle point cloud data that \replaced{are }{is }invariant to spatial transformations.
	
	(3) The JGEKD framework \replaced{was used }{is utilized }for anticorruption training, which effectively enhances the model's robustness against corruption.
	
	\section{Related work}
	
	\subsection{Point cloud task}
	
	In point cloud tasks, multi-view methods \cite{MVCNN, MVTN} or voxel-based methods \cite{Voxnet, 3Dtree, adam2022mesh} describe 3D objects with multiple
	views or \deleted{by }voxelisation, but these methods are demanding in terms of device and scene requirements\deleted{,} and \deleted{also} consume significant memory resources. Point-based methods \cite{P2P, , jiang2023exploiting} can directly operate on point cloud data, extract local features, \deleted{and} are suitable for different objects, \added{and} have become popular \replaced{beacause of }{for }their ease of integration with deep learning. \replaced{Existing studies }{Current work }typically employ\deleted{s} convolutional \cite{PAConv, RS-CNN}, relational \cite{PointASNL}, and graph-based \cite{hu2023vodrac, DGCNN, Grid-GCN} methods to locally aggregate point cloud features, \replaced{however ,}{but }these methods often overlook the internal details of \added{the }point clouds. Hu et al. \cite{hu2020deep} construct\replaced{ed}{s} a feature map from 3D point cloud data, and use\replaced{ed}{s} \added{a} 2D convolution to extract \added{the }features. RepSurf \cite{RepSurf} employ\added{ed} representative surfaces to \deleted{clearly} represent highly localised structures\deleted{,} and \replaced{built}{build} a lightweight network architecture that perform\replaced{ed}{s} well in terms of both performance and speed. PointNeXt \cite{PointNeXt} \replaced{re-examineds }{re-examines }the PointNet++ model and introduce\added{d} residual modules to further improve \added{the }performance. However, the scale of existing point cloud datasets is much smaller than \added{that of }image datasets, \replaced{which }{this }results in poor model generalisation and is not robust to attacks \cite{jing2020self}. To address this issue, Sun et al. \cite{ModelNet-C} propose\added{d} a corruption-robust benchmark for common corrupt point cloud data and employ\added{ed} a fusion sampling data augmentation strategy \cite{Pointmixup, Pointcutmix, RSMix} to achieve robustness. However, these data augmentation strategies assume that \deleted{the }class events are IID, which can result in a lack of inter-class information. In this \replaced{study}{paper}, we examine point cloud data from a non-IID perspective and propose\added{d} \replaced{a }{the }JGEKD strategy for training non-IID point clouds.
	
	\subsection{Knowledge distillation}
	\replaced{The k}{K}nowledge distillation method\added{s} based on feature layers \cite{chen2021distilling} \added{rely }heavily \deleted{relies }on the similarity of network structures. In contrast, the method based on logits is not restricted by network structures and can improve \added{the }model performance without increasing \added{the }computational burden. \deleted{Currently, t}\added{T}hese methods can be categorised into teacher-based distillation \deleted{methods }\cite{KD2015, DKD} and self-distillation methods \cite{shen2022self}. Teacher-based distillation methods require\deleted{s a} powerful network and often exhibits higher performance, \replaced{whereas}{while} self-distillation mechanism\added{s} that transfer\deleted{s} knowledge between different distorted versions of the same training data without relying on accompanying models. In 3D tasks, Zheng et al. \cite{SE-SSD} first introduced the idea of knowledge distillation into point cloud\deleted{s} tasks and presented \replaced{a }{the }Self-Ensembling Single-Stage object Detector (SE-SSD) for outdoor point clouds. Fu et al. \cite{CLpoint} divide\replaced{d}{s} point cloud samples into two parts\replaced{,}{:} local and global, and employ\added{ed} \replaced{a }{the }contrastive distillation method to ensure high consistency in the model's detection of local and global samples. Zhang et al. \cite{PSD} construct\replaced{ed}{s} a Siamese network to \replaced{transfer }{achieve }knowledge transfer for regular point clouds and their perturbed forms. However, these methods neglect inter-class relationships. In this \replaced{study}{paper}, we employ a graph representation to model inter-class relationships and utilize graph distillation strategies to achieve knowledge transfer. \replaced{We }{And }designs two \deleted{different} distillation frameworks based on \added{the }JGEKD strategy, JGEsKD and JGEtKD, to meet the needs of different scenarios.
	
	\subsection{Robustness against corruption of point cloud}
	Xiang et al. \cite{Xiang2019} were the first to provide evidence of the susceptibility of point clouds to adversarial attacks, \replaced{whereas }{while }Zhou et al. \cite{Liu2019} discovered that utilizing input randomization techniques can effectively alleviate this issue. Furthermore, Sun et al. \cite{Sun2021} found that adding pre\deleted{-}training weights to self-supervised tasks can enhance model robustness. \replaced{To }{In order to rigorously }classify the categories of corrupted point clouds\added{ rigorously}, Sun et al. \cite{ModelNet-C} proposed a systematic benchmark and analyzed the corruption robustness of models in the task of point cloud recognition. However, these generalized training methods \replaced{are not pertinent}{lack pertinence}. In this \replaced{study}{paper}, we first apply common corruption transformations \replaced{to }{on }point cloud and employ JGEKD to minimize the distance between the original \deleted{point cloud }and \deleted{the }corrupted point clouds, thereby enhancing the \deleted{model's }robustness\added{ of the model} against corruption.
	
	\section{Our approach}
	In Fig. \ref{ttest}, the point cloud class events follow non-IID, and \deleted{there exist }mutual relationships \added{exist }between \added{the }classes. In Section 3.1, we employ joint graphs to capture \added{the }hidden relationships between classes, calculate the entropy of graphs to represent the amount of information between classes, \added{and }then \replaced{realise }{the }domain migration between classes\deleted{ is realized} by distillation. In Section 3.2, Frameworks JGEsKD and JGEtKD are proposed to achieve knowledge transfer of the same sample in different representations. Finally, in Section 3.3, based on the above framework, an anti-robust training strategy is proposed to \deleted{achieve the learning of }\added{lean} corrupt point clouds.

	\subsection{Joint graph entropy knowledge distillation}
	A graph is defined as $G = (V,E)$ to describe the implicit relationship between classes, the set of nodes is denoted \replaced{as }{by }$V$, \added{which }refers to \replaced{a }{the }set of objects associated with each node. The edges between $N$ nodes \replaced{are }{can be }defined as : $E = \{ {v_i}{v_j}|i = 1,...,N,j = 1,...,N\} $. The output probability of the neural network can be defined as ${\bf{p}} = [{p_1},{p_2},...,{p_i},...,{p_N}] \in {^{1 \times N}}$, and the weight of edge $E$ can be defined as \deleted{follows:}
	\begin{equation}\label{1}
		l({v_i}{v_j}) = {p_i}{p_j},
	\end{equation}
	where ${p_i}{p_j}$ is the joint probability \cite{shannon2001mathematical} between ${v_i}$ and ${v_j}$\replaced{ and represents}{, represent} the implicit relationships between them. We can obtain graphs by matrix multiplication:
	\begin{equation}\label{2}
		{{\bf{A}}} = {{\bf{p}}^{\rm T}}{\bf{p}},
	\end{equation}
	where “$\rm T$” denotes the transpose operation, $\bf{p}$ is the vector with dimension $1 \times N$ , graph ${\bf{A}}$ is the matrix with dimension $N \times N$ , and ${\bf{A}}_{ij}$ represents the joint probability of ${v_i}$ and ${v_j}$ in this sample.\par
	\replaced{Because }{Since }our graph is obtained \replaced{using }{by }the joint probability, we define it as joint graph:
	\begin{equation}\label{3}
		\left\{ {\begin{array}{*{20}{c}}
				{{{\bf{p}}^{\cal S}} = {\cal S}(X)}\\
				{{{\bf{p}}^{\cal T}} = {\cal T}(X)}
		\end{array}} \right.,
	\end{equation}
	where ${{\bf{p}}^{\cal S}}$ and ${{\bf{p}}^{\cal T}}$ represent the predicted probabilities of the output of student network ${\cal S}$ and teacher network ${\cal T}$\added{, respectivly}.
	
	Then, using Eq.(2) to construct joint graphs:
	\begin{equation}\label{4}
		\left\{ {\begin{array}{*{20}{c}}
				{{{\bf{A}}^{\cal S}} = {{({{\bf{p}}^{\cal S}})}^{\rm T}}{{\bf{p}}^{\cal S}}}\\
				{{{\bf{A}}^{\cal T}} = {{({{\bf{p}}^{\cal T}})}^{\rm T}}{{\bf{p}}^{\cal T}}}
		\end{array}} \right..
	\end{equation}
	The value of position $(i,j)$ of\added{ the} matri\replaced{x}{ces} ${{\bf{A}}^{\cal S}}$ and ${{\bf{A}}^{\cal T}}$ can be regarded as the predicted probability and the true probability\added{, respectively}.
	
	When ${\cal S}$ is trained to saturate, we expect ${{\bf{p}}^{\cal S}} \equiv {\bf{p}}{^{\cal T}}$,\added{ where } entropy can be employed to measure how close the predicted probability is to the true probability\replaced{. T}{, then t}he cross entropy of this position can be denoted as $ - {\bf{A}}_{ij}^{\cal S} \times \log ({\bf{A}}_{ij}^{\cal T})$ , where "$\log$" denotes\added{ the} logarithmic function. The joint graph entropy can be defined as
	\begin{equation}\label{5}
		{{\bf{A}}^{\rm{H}}} = -{{\bf{A}}^{\cal S}} \odot \log ({{\bf{A}}^{\cal T}}),
	\end{equation}
	where "${\rm{H}}$" stands for entropy and "$\odot$" denotes \added{the }element-wise multiplication of matrices. \replaced{Then we realized the }{The }knowledge distillation by entropy, and the JGEKD loss function can be defined as \deleted{follows:}
	\begin{equation}\label{6}
		{{\cal L}_{\rm KD}}{\rm{(}}{{\bf{A}}^{\cal T}}\parallel {{\bf{A}}^{\cal S}}{\rm{) = }}\frac{1}{{{N^2}}}{\rm sum}({{\bf{A}}^{\rm{H}}}),
	\end{equation}
	where "$\rm sum$" is matrix summation function: ${\rm{sum(}}{\bf{M)}} = \sum\limits_{i = 1}^N {\sum\limits_{j = 1}^N {\bf{M}} }$.
	
	\replaced{The n}{N}on-IID 3D point clouds exhibit\added{ed} inter-class correlations. Eq\added{s}.(5) and \deleted{Eq.}(6) respectively preserve this information from two perspectives. On the one hand, Eq.(5) describes the inter-class relationship by constructing a joint graph. On the other hand, the ground truth, which is a one-hot encoding with values of 0 or 1, lacks hidden information between classes. In Eq.(6), the probability abstracted by the teacher network is employed instead of the true label, which can guarantee the inter\deleted{-}class correlation to a certain extent, and realiz\replaced{ing}{e} cross-domain transfer across different classes.
	\subsection{Framework knowledge distillation}
	Our knowledge distillation framework is based on \replaced{S}{s}iamese neural networks, which can simultaneously learn different representations of point clouds, and joint graph entropy knowledge distillation, which enables information transfer between different representations of data. The framework for this section comes in two forms: JGEsKD and JGEtKD, which we will cover in turn in Section\added{s} 3.2.1 and \deleted{section }3.2.2.
	
	\subsubsection{Joint graph entropy self knowledge distillation}
	\begin{figure*}[h] 
		\centering 
		\includegraphics[width=\linewidth]{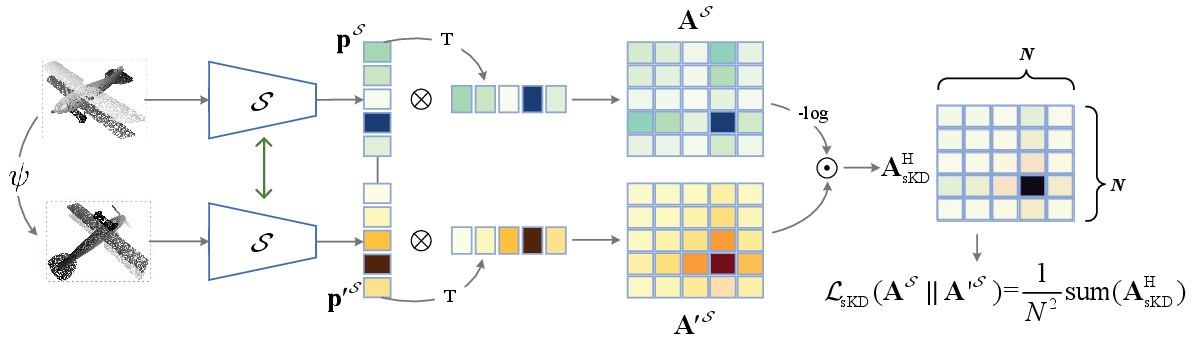}\\
		\caption{Framework of JEsKD.  The \replaced{S}{s}iamese network ${\cal S}$ abstracts the features of the point cloud and its variants to obtain the prediction probabilities ${{\bf{p}}^{\cal S}}$ and ${\bf{p}}^{\prime {\cal S}}$, and then builds the joint graph ${{\bf{A}}^{\cal S}}$ and ${{\bf{A}}^{\prime \cal S}}$ respectively. Finally, the loss was obtained by joint graph entropy knowledge distillation. Here "$\psi$" denotes transformation function, "$\odot$" denotes element-wise multiplication of matrices, "$\otimes$" denotes matrix multiplication, "T" denotes matrix transpose, "log" denotes logarithmic function, and "$N$" denotes number of classes.}
		\label{sKD}
	\end{figure*}
	In Fig. \hypersetup{linkcolor=cyan}\ref{sKD}, we define the initial point cloud data $X$ and its variations $X^\prime=\psi(X)$, and denotes the neural network as ${\cal S}$. The forward propagation process \replaced{is }{can be }defined as follows:
	\begin{equation}\label{7}
		\left\{ {\begin{array}{*{20}{c}}
				{{{\bf{p}}^{\cal S}} = {\cal S}(X)}  \\
				{{\bf{p}}^{\prime S} = {\cal S}(X^\prime )}  \\
		\end{array}} \right.,
	\end{equation}
	where $X$ and $X^\prime $ represent different features of the same sample, therefore, their probabilit\replaced{ies}{y} abstracted through ${\cal S}$ should satisfy\deleted{:} ${{\bf{p}}^{\cal S}} \equiv {\bf{p}}{^{\prime \cal S}}$. \par 
	Therefore, combining ${{\bf{p}}^{\cal S}}$ and ${\bf{p}}^{\prime {\cal S}}$ to supervise the network model can improve the generalization ability of the model. Combine Eq\added{s}.(5) and \deleted{Eq.}(8), the joint graph can be defined as\deleted{:}
	\begin{equation}\label{8}
		\left\{ {\begin{array}{*{20}{c}}
				{{{\bf{A}}^{\cal S}}{\bf{ = }}{{({{\bf{p}}^{\cal S}})}^{\rm T}}{{\bf{p}}^{\cal S}}}\\
				{{\bf{A}}{^{\prime \cal S}}{\bf{ = }}{{({\bf{p}}{^{\prime \cal S}})}^{\rm T}}{\bf{p}}{^{\prime \cal S}}}
		\end{array}} \right..
	\end{equation}
	
	Combin\replaced{e}{ing} Eq\added{s}.(5) and \deleted{Eq.}(8), the joint entropy matrix of \replaced{sKD}{self knowledge distillation} is defined as:
	\begin{equation}\label{9}
		{\bf{A}}_{{\rm{sKD}}}^{\rm{H}}{\bf{ = }}{-{\bf{A}}^{\cal S}} \odot \log{\bf{(A}}{^{\prime \cal S}}{\bf{)}}.
	\end{equation}
	
	Then, combining Eq\added{s}.(6) and \deleted{Eq.}(9), joint graph entropy self knowledge distillation (JGEsKD) loss function can be defined as \deleted{follows:}
	\begin{equation}\label{10}
		{{\cal L}_{{\rm{sKD}}}}{\rm{(}}{{\bf{A}}^{\cal S}}\parallel {{\bf{A}}{^{\prime \cal S}}}{\rm{) = }}\frac{1}{{{N^2}}}{\rm sum}({\bf{A}}_{{\rm{sKD}}}^{\rm{H}})
	\end{equation}
	
	\subsubsection{Joint graph entropy teacher knowledge distillation}
	\begin{figure*}[t]
		\centering 
		\includegraphics[width=\linewidth]{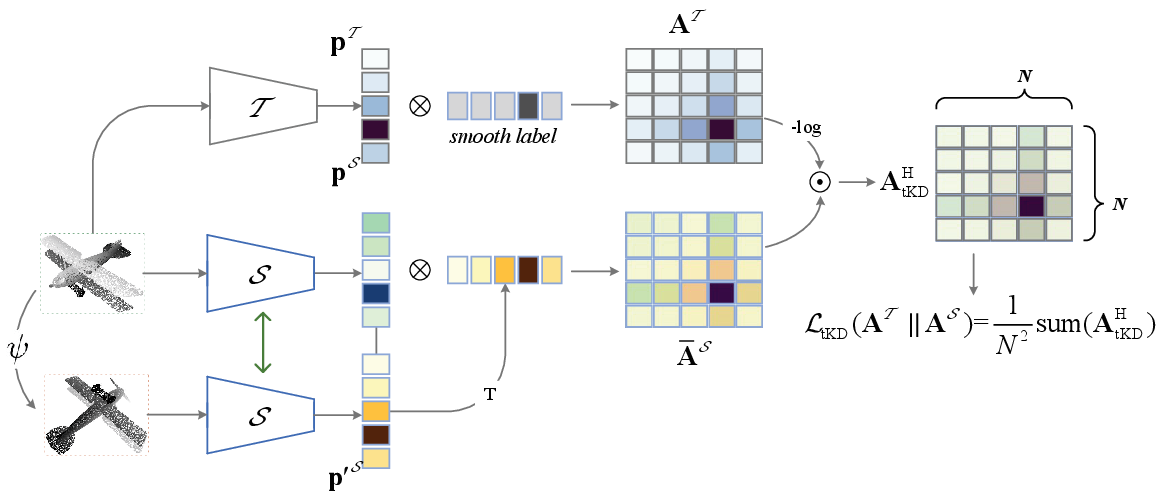}\\
		\caption{Framework of JGEtKD. The teacher ${\cal T}$ abstracts the features of the point cloud and the student network ${\cal S}$ abstracts the features of the point cloud and its variants, and then builds the joint graph ${{\bf{A}}^{\cal T}}$ and ${{\bf{\bar A}}^{\cal S}}$\deleted{ respectively}. Finally, the loss was obtained by joint graph entropy knowledge distillation. Here "$\psi$", "$\odot$", "$\otimes$", "T", "log" and "$N$" is the same as Fig. 2.}
		\label{tKD}
	\end{figure*}
	
	Fig. \hypersetup{linkcolor=cyan}\ref{tKD} \replaced{illustrates }{shows }the JGEtKD framework. The forward propagation of ${\cal S}$ is\replaced{ identical to that of}{ the same as} Eq.(7); \replaced{however }{but }there \replaced{were }{are }differences in the construction of\added{ the} joint graph. 
	
	The joint graph of \added{the }student network output under the framework of JGEtKD can be defined as
	\begin{equation}\label{11}
		{{\bf{\bar A}}^{\cal S}}{\bf{ = }}{({{\bf{p}}^{\cal S}})^{\rm T}}{\bf{p}}{^{\prime \cal S}},
	\end{equation}
	\replaced{this }{which }can be understood as \replaced{an }{the }exchange of hidden class information \replaced{for }{of }the same sample under different representations.

	In addition, we need to define \added{the }teacher network ${\cal T}$ and \added{the }ground truth $q = [{q_1},{q_2},...,{q_i},...,{q_N}] \in {\mathcal{R}^{1 \times N}}$. The prediction probability of \added{the} teacher network ${\cal T}$ output is ${{\bf{p}}^{\cal T}} = {\cal T}(X)$. The joint graph can be defined as: ${{\bf{A}}^{\cal T}}{\bf{ = q}}^\prime {{\bf{p}}^{\cal T}}$ , where ${\bf{q}}^\prime $ is the smooth label: $q^\prime = q \times (1 - \varepsilon ) + (1 - q) \times \varepsilon /({\rm N} - 1)$, and $\varepsilon $ is the smoothing ratio. \replaced{This }{It }can effectively preserve \deleted{the }category information in ${{\bf{p}}^{\cal T}}$ and avoid excessive zeros in ${{\bf{A}}^{\cal T}}$. 
	
	We \added{in}put ${{\bf{\bar A}}^{\cal S}}$ and ${{\bf{A}}^{\cal T}}$ into Eq.(5) to obtain the joint graph entropy:
	\begin{equation}\label{12}
		{\bf{A}}_{{\rm{tKD}}}^{\rm{H}}{\bf{ = -}}{{\bf{\bar A}}^{\cal S}} \odot \log{\bf{(}}{{\bf{A}}^{\cal T}}{\bf{)}}.
	\end{equation}
	
	Then combining Eq\added{s}.(6) and \deleted{Eq.}(12), the JEGtKD loss \replaced{for }{of }a single sample can be defined as\deleted{ follows:}
	\begin{equation}\label{13}
		{{\cal L}_{{\rm{tKD}}}}{\rm{(}}{{\bf{A}}^{\cal T}}\parallel {{\bf{A}}^{\cal S}}{\rm{) = }}\frac{1}{{{N^2}}}{\rm sum}({\bf{A}}_{{\rm{tKD}}}^{\rm{H}})
	\end{equation}
	
	\replaced{Theoretically}{In theory}, the teacher-knowledge distillation framework based on a high-performance teacher model is expected to outperform the self-knowledge distillation framework. However, \replaced{a }{the }self-knowledge framework \replaced{incurs}{have} lower training costs.
	
	\subsection{Robustness against corruption training}
	Neural networks are not robust to corrupt point clouds because they do not learn knowledge of corrupt point clouds. To facilitate a holistic understanding of point cloud morphology while training networks to identify corrupted data, this \replaced{study }{paper }proposes a robust training strategy against corruption.
	
	Based on the framework in Section 3.2, we build a Siamese neural network, simultaneously abstract the features of \added{the} conventional point cloud $X$ and its corrupted form $X^\prime$, and realize their knowledge transfer through\deleted{ the method of} knowledge distillation. The corruption point cloud $X^\prime$ can be defined as follows:
	\begin{equation}\label{14}
		X^\prime = {\rm c}(\psi _k^d,\psi _0^o) \circ {\rm c}(\psi _j^n,\psi _0^o) \circ \psi _i^t(X),
	\end{equation}
	where $\psi$ denotes\added{ the} transformation function,  the superscripts $t$, $n$, and $d$ represent morphological transformations (includes Rotation, Shear, free-form deformation (FFD), radial basis function (RBF), Lnv\_RBF), noise (includes Gaussian, Impulse, Uniform, Upsampling, Background), and density (includes Occlusion, Light Detection And Ranging (LiDAR), Density\_lnc., Density\_Dec., Cutout), respectively. The subscript indicates the number of \deleted{the }specific transformation\added{s}, where $i \in [0,5), j \in [0,4), k \in [0,3)$ and the \deleted{specific }corresponding \replaced{relationships are presented }{relationship is shown }in Table \ref{tab 1}. We refer to ${\mathop{\rm c}\nolimits}$ as \replaced{the selection }{selecting }of functions: ${\mathop{\rm c}\nolimits} (f,g) = f \vee g$, and "$\circ$" as \added{the }nesting of functions: $f \circ g(x) = f(g(x))$, and $\psi _0^o$ is the identity function, which denotes that no transformation takes place: $\psi _0^o(X) = X$.

	\begin{table}[!ht]
		\centering
		\caption{Corrupt transformations and their function definition.}\label{tab 1}
		\begin{tabular}{cc|cc|cc}
			\hline
			\multicolumn{2}{c|}{Transformation}         & \multicolumn{2}{c|}{Noise}                         & \multicolumn{2}{c}{Density}             \\ \hline
			\multicolumn{1}{c|}{Rotation}  & $\psi _0^t$  & \multicolumn{1}{c|}{Gaussian}   & $\psi _0^n$  & \multicolumn{1}{c|}{Occlusion}     & -           \\
			\multicolumn{1}{c|}{Shear}     & $\psi _1^t$  & \multicolumn{1}{c|}{Impulse}    & $\psi _1^n$  & \multicolumn{1}{c|}{LiDAR}         & -           \\
			\multicolumn{1}{c|}{FFD}       & $\psi _2^t$  & \multicolumn{1}{c|}{Uniform}    & $\psi _2^n$  & \multicolumn{1}{c|}{Density\_lnc.} & $\psi _0^d$ \\
			\multicolumn{1}{c|}{RBF}       & $\psi _3^t$  & \multicolumn{1}{c|}{Upsampling} & $\psi _3^n$  & \multicolumn{1}{c|}{Density\_Dec.} & $\psi _1^d$ \\
			\multicolumn{1}{c|}{Lnv\_RBF}  & $\psi _4^t$  & \multicolumn{1}{c|}{Background} & -            & \multicolumn{1}{c|}{Cutout}        & $\psi _2^d$ \\ \hline
		\end{tabular}
		\justify
		\footnotesize 'Background' alters the number of points in \replaced{a }{the }sample, resulting in an inability to concatenate \added{the }samples during batch processing. 'Occlusion' and 'LiDAR' generate corrupted point clouds through scanning of the target, rather than direct manipulation of point-based data.
	\end{table}
	
	We employed \replaced{a random data augmentation strategy}{the strategy of random data augmentation}, whereby each sample \replaced{underwent }{undergoes }distinct transformations in each epoch during training. In principle, with sufficient training iterations, the neural network can learn the knowledge of all corrupted transformations for each sample.

	It is noteworthy that, using the JGEsKD framework as an example, corrupt point clouds can be expressed as $X^\prime = \psi (X)$, some point cloud data is corrupt and lacks essential information. As per Eq.(7), the neural network abstracts the features of $X$ and $X^\prime$ to obtain ${{\bf{p}}^{\cal S}}$ and ${\bf{p}}^{\prime \cal S}$, respectively. It can be challenging to minimize the distance between ${{\bf{p}}^{\cal S}}$ and ${\bf{p}}^{\prime \cal S}$ via knowledge distillation, as it can lead to incorrect information in the loss function during the solving process. To mitigate this issue, we incorporate both knowledge distillation loss and label loss in the training process, and the overall loss function is defined as follows:
	\begin{equation}\label{15}
		{\cal L} = \alpha {{\cal L}_{\rm CE}} + \beta {{\cal L}_{\rm KD}},
	\end{equation}
	where ${{\cal L}_{\rm CE}}$ is the cross entropy loss function, ${{\cal L}_{\rm KD}}$ is the knowledge distillation loss function, and $\alpha$ and $\beta$ are the proportional coefficients of cross entropy \deleted{loss }and distillation loss respectively.
	\section{Experiments}

	We conducted \replaced{comparative }{comparison }experiments on three datasets to evaluate the overall effectiveness of the approach. Furthermore, \deleted{we performed }ablation experiments\added{ were performed} to assess the effectiveness of the JGEKD loss function and \deleted{the }framework. Additionally, we \deleted{also }conducted a corruption robustness test on ModelNet40-C to evaluate \replaced{its }{the }robustness against corruption\deleted{ of the model}.
	\subsection{Setup}
	\subsubsection{Dataset}
	(1) Real-world object dataset

	ScanObjectNN \cite{ScanObject}: This is a dataset of real point cloud objects based on indoor scans, containing 15 classes and approximately 15,000 samples. We adopted \replaced{the }{its }hardest variant (PB\_T50\_RS\deleted{ variant}) as the training object to evaluate the effectiveness of our proposed strategy.
	
	ScanntV2\_cls \cite{Scannet}: ScanNet is a dataset of 3D indoor scenes that have been\added{ extensively} reconstructed and annotated\deleted{ extensively}. Following the methodology outlined in \cite{PointGLR}, we employ\added{ed} \deleted{the }ScanNetV2 annotations and splits, which contain 1,201 scenes for training and 312 scenes for testing across 17 classes. For the classification task, we selected 12,060 objects for training and 3,416 objects for testing.
	
	(2) Human-made object dataset
	
	ModelNet40 \cite{ModelNet40}: This dataset features diverse categories, clear shapes, and a well-structured organization. It consists of 40 classes and \deleted{contains }12,311 samples, with 9,843 samples employed for training and 2,468 samples employed for testing.
	
	ModelNet40-C \cite{ModelNet-C}: Sun et al. introduced three types of corruptions, “Transformation”, “Noise”, and “Density”, to the dataset and generated the robust  corruption dataset ModelNet40-C. Each type of corruption has five benchmarks, and each benchmark is divided into five levels. As the level increases, the degree of point cloud data corruption and task difficulty also increase\deleted{s}.
	
	\subsubsection{Networks}
	\replaced{Because}{Since} JGEKD is a general training strategy that is agnostic \replaced{to }{of }the network, architectures employed, we chose four 3D visual networks including the classic model\added{s} PointNet++$\dag$ \cite{Pointnet++}, RepSurf-U$\ddag$ \cite{RepSurf}, RepSurf-U$\ddag$2X \cite{RepSurf}, and PointNetXt-S \cite{PointNeXt} to evaluate its performance.

	\subsubsection{Implementation details}

	All experiments conducted in this study were performed on a Tesla V100S-PCIE-32GB GPU employing the Pytorch \cite{pytorch} framework. The fraction\added{s} of \added{the }label \deleted{loss }and distillation losses during the distillation process was maintained at a 1:1 ratio. To ensure the fairness of the experiments, all other hyperparameters employed in the distillation training process were fully consistent with the benchmark networks employed. It is noteworthy that The teacher network in JGEtKD framework adopts JGEsKD pretraining weight, whereas the pre-trained model was not utilized in the other strategy. The \deleted{evaluation of }the different strategies \replaced{were evaluated }{was }based on \added{the }overall accuracy (OA, \%) and mean per-class accuracy (mAcc, \%).
	\subsection{Comparative experiment}
	\begin{table*}[!t]
		\centering
		\caption{Performance of classification on ScanObjectNN, ModelNet40 and ScanntV2\_cls.}\label{tab 2}
		\resizebox{\linewidth}{!}{%
		\begin{tabular}{llrrrrrrrrrrrrrrrr}   
			\hline
			\multicolumn{1}{l}{}                    & \multicolumn{1}{l}{} & \textbf{Rotation} & \textbf{Shear}    & \textbf{FFD}      & \textbf{RBF}      & \textbf{Lnv.RBF}  & \textbf{Uniform}  & \textbf{Ups.}     & \textbf{Gaussian} & \textbf{Impulse}  & \textbf{Bg.}      & \textbf{Occlusion} & \textbf{LiDAR}    & \textbf{Inc.}     & \textbf{Dec.}     & \textbf{Cutout}   & \textbf{m\_CE}    \\ \hline
			\multirow{6}{*}{\textbf{PointNet++}}     & \textbf{ST}          & 1.0(2)            & 1.0(4)            & 1.0(4)            & 1.0(5)            & 1.0(5)            & 1.0(5)            & 1.0(6)            & 1.0(5)            & 1.0(6)            & 1.0(4)            & 1.0(6)             & 1.0(6)            & 1.0(6)            & 1.0(6)            & 1.0(6)            & 1.0(5)            \\
			& \textbf{cutmix\_k}   & 1.04(3)           & 0.894(2)          & 0.913(2)          & 0.857(2)          & 0.902(2)          & 0.982(4)          & 0.906(2)          & 0.977(4)          & 0.46(4)           & 0.799(3)          & 0.876(3)           & 0.796(3)          & 0.406(4)          & 0.511(2)          & \textbf{0.569(1)} & 0.793(2)          \\
			& \textbf{cutmix\_r}   & 1.085(5)          & 1.025(5)          & 1.042(5)          & 0.879(4)          & 0.936(4)          & 0.838(2)          & 0.962(5)          & 0.744(2)          & 0.358(2)          & 1.14(5)           & 0.953(4)           & 0.788(2)          & 0.391(2)          & 0.512(3)          & 0.6(2)            & 0.817(3)          \\
			& \textbf{mixup}       & 1.068(4)          & 0.923(3)          & 0.955(3)          & 0.872(3)          & 0.905(3)          & 0.912(3)          & 0.917(3)          & 0.868(3)          & 0.55(5)           & 1.751(6)          & 0.964(5)           & 0.996(5)          & 0.562(5)          & 0.747(5)          & 0.751(5)          & 0.916(4)          \\
			& \textbf{rsmix}       & 1.819(6)          & 1.305(6)          & 1.45(6)           & 1.562(6)          & 1.604(6)          & 1.897(6)          & 0.919(4)          & 2.098(6)          & 0.433(3)          & 0.725(2)          & 0.82(2)            & 0.799(4)          & 0.404(3)          & 0.524(4)          & 0.612(4)          & 1.131(6)          \\
			& \textbf{DesenAT-sD}  & \textbf{0.755(1)} & \textbf{0.555(1)} & \textbf{0.808(1)} & \textbf{0.733(1)} & \textbf{0.765(1)} & \textbf{0.677(1)} & \textbf{0.866(1)} & \textbf{0.586(1)} & \textbf{0.331(1)} & \textbf{0.52(1)}  & \textbf{0.808(1)}  & \textbf{0.673(1)} & \textbf{0.38(1)}  & \textbf{0.494(1)} & 0.602(3)          & \textbf{0.637(1)} \\ \hline
			\multirow{6}{*}{\textbf{PointNetMeta-s}} & \textbf{ST}          & 0.961(2)          & 0.922(3)          & 0.961(4)          & 0.948(4)          & 0.999(4)          & 0.952(3)          & 0.955(6)          & 0.922(3)          & 1.228(6)          & 0.881(5)          & 0.867(3)           & 0.934(6)          & 0.426(6)          & 0.902(6)          & 0.923(6)          & 0.919(5)          \\
			& \textbf{cutmix\_k}   & 1.266(5)          & 0.998(5)          & 1.019(5)          & 1.018(5)          & 1.058(5)          & 1.598(5)          & 0.811(2)          & 1.7(5)            & 0.436(4)          & 0.664(2)          & \textbf{0.778(1)}  & 0.738(3)          & 0.315(3)          & 0.498(3)          & \textbf{0.539(1)} & 0.896(4)          \\
			& \textbf{cutmix\_r}   & 1.092(4)          & 0.942(4)          & 0.956(3)          & 0.898(3)          & 0.912(3)          & 0.997(4)          & \textbf{0.799(1)} & 0.931(4)          & \textbf{0.317(1)} & 0.747(3)          & 0.913(6)           & \textbf{0.725(1)} & \textbf{0.301(1)} & \textbf{0.476(1)} & 0.553(3)          & 0.771(2)          \\
			& \textbf{mixup}       & 0.999(3)          & 0.829(2)          & 0.897(2)          & 0.848(2)          & 0.885(2)          & 0.92(2)           & 0.892(5)          & 0.84(2)           & 0.445(5)          & 1.377(6)          & 0.901(4)           & 0.899(5)          & 0.38(5)           & 0.662(5)          & 0.676(5)          & 0.83(3)           \\
			& \textbf{rsmix}       & 1.606(6)          & 1.071(6)          & 1.238(6)          & 1.347(6)          & 1.376(6)          & 2.034(6)          & 0.82(3)           & 2.148(6)          & 0.342(3)          & 0.791(4)          & 0.903(5)           & 0.736(2)          & 0.313(2)          & 0.477(2)          & 0.539(2)          & 1.049(6)          \\
			& \textbf{DesenAT-sD}  & \textbf{0.939(1)} & \textbf{0.556(1)} & \textbf{0.896(1)} & \textbf{0.813(1)} & \textbf{0.856(1)} & \textbf{0.671(1)} & 0.87(4)           & \textbf{0.565(1)} & 0.322(2)          & \textbf{0.457(1)} & 0.807(2)           & 0.8(4)            & 0.35(4)           & 0.558(4)          & 0.602(4)          & \textbf{0.671(1)} \\ \hline
			\multirow{6}{*}{\textbf{APES\_global}}   & \textbf{ST}          & 0.784(3)          & 0.673(2)          & 0.792(3)          & 0.878(5)          & 0.905(5)          & 0.892(4)          & 0.852(3)          & 0.873(5)          & 0.711(6)          & 0.444(3)          & 0.973(3)           & 0.98(6)           & 0.389(4)          & 0.709(6)          & 0.828(6)          & 0.779(4)          \\
			& \textbf{cutmix\_k}   & 0.913(6)          & 0.86(6)           & 0.983(6)          & 0.92(6)           & 0.964(6)          & 1.004(6)          & 1.0(6)            & 0.946(6)          & 0.545(4)          & 0.684(5)          & 1.054(6)           & 0.934(5)          & 0.449(6)          & 0.703(5)          & 0.785(3)          & 0.85(6)           \\
			& \textbf{cutmix\_r}   & 0.865(5)          & 0.837(5)          & 0.93(5)           & 0.835(3)          & 0.873(3)          & 0.809(2)          & 0.894(4)          & 0.752(2)          & 0.436(2)          & 0.563(4)          & 0.979(4)           & 0.887(4)          & 0.361(3)          & 0.579(3)          & 0.827(5)          & 0.762(3)          \\
			& \textbf{mixup}       & 0.755(2)          & 0.764(4)          & 0.878(4)          & 0.845(4)          & 0.884(4)          & 0.9(5)            & 0.983(5)          & 0.832(3)          & 0.605(5)          & 0.9(6)            & 1.031(5)           & 0.878(3)          & 0.389(5)          & 0.691(4)          & 0.807(4)          & 0.809(5)          \\
			& \textbf{rsmix}       & 0.823(4)          & 0.684(3)          & 0.758(2)          & 0.762(2)          & 0.811(2)          & 0.882(3)          & 0.837(2)          & 0.85(4)           & 0.506(3)          & 0.441(2)          & 0.92(2)            & \textbf{0.842(1)} & \textbf{0.303(1)} & \textbf{0.462(1)} & \textbf{0.53(1)}  & 0.694(2)          \\
			& \textbf{DesenAT-sD}  & \textbf{0.619(1)} & \textbf{0.497(1)} & \textbf{0.677(1)} & \textbf{0.689(1)} & \textbf{0.723(1)} & \textbf{0.603(1)} & \textbf{0.778(1)} & \textbf{0.529(1)} & \textbf{0.322(1)} & \textbf{0.369(1)} & \textbf{0.903(1)}  & 0.856(2)          & 0.314(2)          & 0.539(2)          & 0.649(2)          & \textbf{0.604(1)} \\ \hline
			\multirow{6}{*}{\textbf{APES\_local}}    & \textbf{ST}          & 0.757(3)          & 0.708(4)          & 0.78(4)           & 0.847(4)          & 0.873(4)          & 0.87(4)           & 0.879(4)          & 0.831(4)          & 0.631(4)          & 0.42(2)           & 0.948(2)           & 0.921(6)          & 0.383(4)          & 0.7(4)            & 0.799(4)          & 0.756(4)          \\
			& \textbf{cutmix\_k}   & 0.984(6)          & 1.011(5)          & 1.161(6)          & 1.082(6)          & 1.115(6)          & 1.217(6)          & 1.246(6)          & 1.128(6)          & 0.703(6)          & 1.066(6)          & 0.997(5)           & 0.869(4)          & 0.492(6)          & 0.807(6)          & 0.876(5)          & 0.984(6)          \\
			& \textbf{cutmix\_r}   & 0.96(5)           & 1.048(6)          & 1.07(5)           & 1.005(5)          & 1.069(5)          & 1.044(5)          & 1.215(5)          & 0.974(5)          & 0.642(5)          & 0.982(5)          & 1.062(6)           & 0.88(5)           & 0.443(5)          & 0.732(5)          & 0.876(6)          & 0.933(5)          \\
			& \textbf{mixup}       & \textbf{0.656(1)} & 0.645(2)          & 0.756(2)          & 0.736(2)          & 0.761(2)          & 0.757(2)          & 0.875(3)          & 0.668(2)          & 0.474(3)          & 0.734(4)          & 0.959(4)           & 0.854(3)          & 0.353(3)          & 0.617(3)          & 0.655(2)          & 0.7(3)            \\
			& \textbf{rsmix}       & 0.848(4)          & 0.675(3)          & 0.758(3)          & 0.745(3)          & 0.774(3)          & 0.862(3)          & 0.843(2)          & 0.809(3)          & 0.43(2)           & 0.446(3)          & 0.957(3)           & \textbf{0.835(1)} & \textbf{0.303(1)} & \textbf{0.467(1)} & \textbf{0.536(1)} & 0.686(2)          \\
			& \textbf{DesenAT-sD}  & 0.709(2)          & \textbf{0.52(1)}  & \textbf{0.718(1)} & \textbf{0.696(1)} & \textbf{0.716(1)} & \textbf{0.65(1)}  & \textbf{0.813(1)} & \textbf{0.576(1)} & \textbf{0.373(1)} & \textbf{0.391(1)} & \textbf{0.925(1)}  & 0.853(2)          & 0.339(2)          & 0.573(2)          & 0.674(3)          & \textbf{0.635(1)} \\ \hline
		\end{tabular}
		}
		\justifying
		\footnotesize \dag: single-scale grouping (SSG), \ddag: multi-scale inference from \cite{RepSurf}, 2X: with double channels, ST: standard training. 
	\end{table*}
	In Table \hypersetup{linkcolor=cyan}\ref{tab 2}, we first\deleted{ly} compare JGEsKD and JGEtKD with the current advanced algorithms, and evaluate \added{the }different strategies in terms of OA and mAcc. Next, we conducted \deleted{a }network weights analysis to reveal the training differences between JGEtKD and standard training (ST), explain\added{ing} the advantages of the training method \deleted{in this paper }from the perspective of network weights.

	% \usepackage{multirow}
	\begin{table*}[!t]
	\centering
	\caption{Performance of classification on ScanObjectNN, ModelNet40 and ScanntV2\_cls.}\label{tab 2}
	\resizebox{\linewidth}{!}{%
		\begin{tabular}{llrrrrrrrr}
			\hline
			\multicolumn{1}{l}{}                    &                     & \textbf{add\_global} & \textbf{add\_local} & \textbf{dropout\_global} & \textbf{dropout\_local} & \textbf{jitter}   & \textbf{rotate}   & \textbf{scale}    & \textbf{m\_CE}    \\ \hline
			\multirow{6}{*}{\textbf{PointNet++}}     & \textbf{ST}         & 1.0(6)               & 1.0(6)              & 1.0(5)                   & 1.0(6)                  & 1.0(4)            & 1.0(2)            & 1.0(6)            & 1.0(5)            \\
			& \textbf{cutmix\_k}  & 0.921(4)             & 0.896(4)            & 0.592(3)                 & 0.458(2)                & 1.019(5)          & 1.09(4)           & 0.953(4)          & 0.847(3)          \\
			& \textbf{cutmix\_r}  & 0.961(5)             & 0.942(5)            & 0.373(2)                 & 0.629(3)                & 0.598(2)          & 1.129(5)          & 0.983(5)          & 0.802(2)          \\
			& \textbf{mixup}      & 0.893(2)             & 0.875(3)            & 1.022(6)                 & 0.767(5)                & 0.945(3)          & 1.05(3)           & 0.832(3)          & 0.912(4)          \\
			& \textbf{rsmix}      & \textbf{0.89(1)}     & \textbf{0.829(1)}   & 0.736(4)                 & \textbf{0.454(1)}       & 1.973(6)          & 1.552(6)          & 0.796(2)          & 1.033(6)          \\
			& \textbf{DesenAT-sD} & 0.919(3)             & 0.859(2)            & \textbf{0.368(1)}        & 0.642(4)                & \textbf{0.385(1)} & \textbf{0.862(1)} & \textbf{0.616(1)} & \textbf{0.664(1)} \\ \hline
			\multirow{6}{*}{\textbf{PointNetMeta-s}} & \textbf{ST}         & 1.009(6)             & 0.941(6)            & 1.208(6)                 & 0.591(6)                & 1.081(4)          & 1.006(2)          & 1.022(6)          & 0.98(6)           \\
			& \textbf{cutmix\_k}  & \textbf{0.798(1)}    & \textbf{0.79(1)}    & 0.839(4)                 & 0.278(2)                & 1.783(5)          & 1.195(5)          & 0.864(4)          & 0.935(4)          \\
			& \textbf{cutmix\_r}  & 0.882(5)             & 0.814(3)            & 0.516(2)                 & 0.383(3)                & 0.901(3)          & 1.12(4)           & 0.918(5)          & 0.791(2)          \\
			& \textbf{mixup}      & 0.863(4)             & 0.847(5)            & 0.932(5)                 & 0.432(4)                & 0.819(2)          & 1.033(3)          & 0.794(2)          & 0.817(3)          \\
			& \textbf{rsmix}      & 0.841(2)             & 0.809(2)            & 0.65(3)                  & \textbf{0.241(1)}       & 2.029(6)          & 1.4(6)            & 0.806(3)          & 0.968(5)          \\
			& \textbf{DesenAT-sD} & 0.86(3)              & 0.838(4)            & \textbf{0.477(1)}        & 0.478(5)                & \textbf{0.359(1)} & \textbf{0.953(1)} & \textbf{0.525(1)} & \textbf{0.641(1)} \\ \hline
			\multirow{6}{*}{\textbf{APES\_global}}   & \textbf{ST}         & 0.784(2)             & 0.756(2)            & 0.706(4)                 & 0.457(4)                & 0.87(4)           & 0.802(3)          & 0.811(2)          & 0.741(4)          \\
			& \textbf{cutmix\_k}  & 0.956(5)             & 0.923(5)            & 0.929(6)                 & 0.4(2)                  & 0.902(5)          & 0.912(6)          & 0.974(6)          & 0.857(6)          \\
			& \textbf{cutmix\_r}  & 0.86(4)              & 0.861(4)            & 0.388(2)                 & 0.499(5)                & 0.583(2)          & 0.871(4)          & 0.916(4)          & 0.711(3)          \\
			& \textbf{mixup}      & 1.0(6)               & 0.971(6)            & 0.872(5)                 & 0.533(6)                & 0.708(3)          & 0.762(2)          & 0.945(5)          & 0.827(5)          \\
			& \textbf{rsmix}      & 0.835(3)             & 0.799(3)            & \textbf{0.359(1)}        & \textbf{0.256(1)}       & 0.911(6)          & 0.875(5)          & 0.834(3)          & 0.696(2)          \\
			& \textbf{DesenAT-sD} & \textbf{0.78(1)}     & \textbf{0.736(1)}   & 0.427(3)                 & 0.421(3)                & \textbf{0.347(1)} & \textbf{0.69(1)}  & \textbf{0.519(1)} & \textbf{0.56(1)}  \\ \hline
			\multirow{6}{*}{\textbf{APES\_local}}    & \textbf{ST}         & 0.855(3)             & 0.817(3)            & 0.692(5)                 & 0.455(4)                & 0.796(4)          & 0.762(3)          & 0.779(2)          & 0.737(4)          \\
			& \textbf{cutmix\_k}  & 1.225(6)             & 1.206(5)            & 0.948(6)                 & 0.485(5)                & 1.037(6)          & 0.988(6)          & 1.152(6)          & 1.006(6)          \\
			& \textbf{cutmix\_r}  & 1.191(5)             & 1.212(6)            & 0.638(3)                 & 0.572(6)                & 0.776(3)          & 0.945(5)          & 1.123(5)          & 0.922(5)          \\
			& \textbf{mixup}      & 0.862(4)             & 0.835(4)            & 0.648(4)                 & 0.413(2)                & 0.512(2)          & \textbf{0.675(1)} & 0.824(3)          & 0.681(2)          \\
			& \textbf{rsmix}      & 0.836(2)             & 0.789(2)            & \textbf{0.382(1)}        & \textbf{0.266(1)}       & 0.808(5)          & 0.894(4)          & 0.832(4)          & 0.687(3)          \\
			& \textbf{DesenAT-sD} & \textbf{0.795(1)}    & \textbf{0.754(1)}   & 0.503(2)                 & 0.422(3)                & \textbf{0.404(1)} & 0.749(2)          & \textbf{0.593(1)} & \textbf{0.603(1)} \\ \hline
		\end{tabular}
		}
		\justifying
		\footnotesize \dag: single-scale grouping (SSG), \ddag: multi-scale inference from \cite{RepSurf}, 2X: with double channels, ST: standard training. 
		\end{table*}
	\subsubsection{Result analysis}
	First, \deleted{we analyze }the results \replaced{were analysed using }{under }the OA evaluation metric. \replaced{For }{On }ScanObjectNN with background interference, PointNet++$\dag$ was improved by 6.87\% and 7.25\%\deleted{, respectively,} by using\added{ the} JGEsKD and JGEtKD strategies\added{, respectively}. RepSurf-U$\ddag$ \replaced{improved by }{An increase of }2.71\% and 3.33\%\deleted{respectively} compared \replaced{with }{to }the baseline. RepSurf-U$\ddag$2X \replaced{improved by }{is }3.22\% and 3.99\%\deleted{ respectively} compared to the baseline. PointNetXt-s \replaced{improved by }{is }1.18\% and 1.45\% \deleted{respectively }compared \replaced{with }{to }the baseline. Although the relationships between classes in the manually processed dataset are not as tight as the real-world scenario dataset ScanObject, our strategy \deleted{is }still improved relative to the baseline, RepSurf-U$\ddag$2X by 0.2\% and 0.44\% in ModelNet40 and 0.44\% and 0.59\% in ScanntV2\_cls.
	
	\replaced{We t}{T}hen analyz\replaced{sed}{ing} the results \replaced{using }{under }the mAcc evaluation metric. \replaced{This }{Here }is an interesting phenomenon: the numerical difference in OA and mAcc obtained by our strategies (JGEsKD and JGEtKD) compared \replaced{with }{to }ST is much smaller. This is because our strategy \deleted{can }preserve\added{s} the implicit relationship between classes and \replaced{stabilises}{make }their prediction\deleted{ between classes more stable}. Using the ScanObjectNN dataset as an example, PointNet++$\dag$, RepSurf-U$\ddag$, RepSurf-U$\ddag$2X and PointNeXt-S under the ST strategy, the numerical differences between the OA and mAcc are 2.5\%, 2.7\%, 2.92\%, and 3.18\%. However, the numerical difference\added{s} between \added{the }OA and mAcc of JGEtKD strategy \replaced{were }{was }only 1.51\%, 0.90\%, 0.66\% and 2.65\%, indicating that our strategy was more stable across classes.
	\subsubsection{Weight and training process analysis}
	\begin{figure*}[!ht] 
		\centering 
		\includegraphics[width=0.7\linewidth]{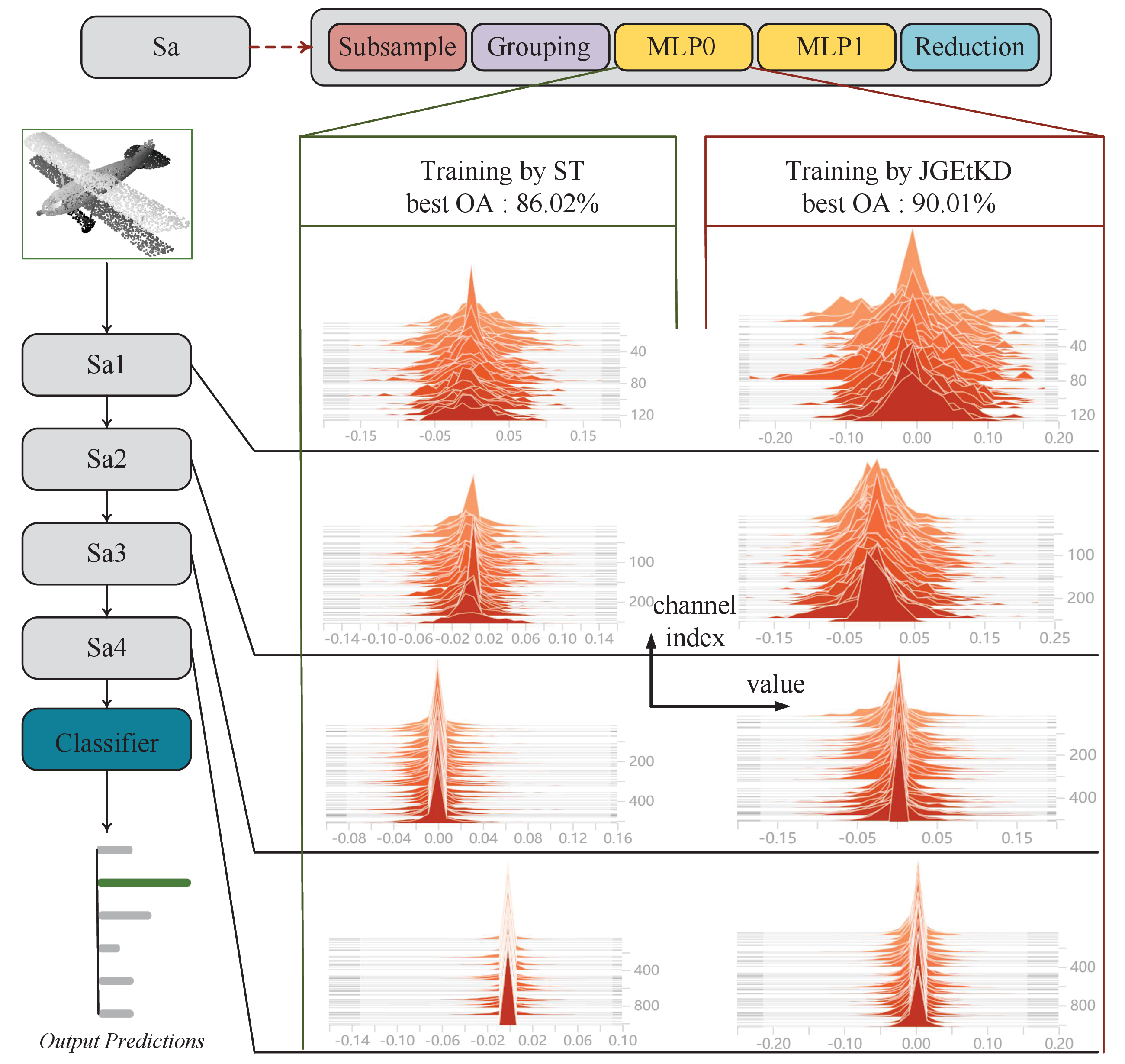}\\
		\caption{Training process and weight analysis plot: The left side of the figure is an overview of the input flow of RepSurf-U$\ddag$2X for classification, and the right side is the weight histogram of "MLP0" in each set
			abstraction (Sa) layer.}
		\label{W}
	\end{figure*}

	In Fig. \hypersetup{linkcolor=cyan}\ref{W}, it can be intuitively seen that JGEtKD can achieve\added{ a} better fitting effect in the early stage of training compared with ST. The shallow weight (Sa1, Sa2) of the model trained by JGEtKD is more discrete than that obtained by ST, which makes our model more effective\replaced{, and}{:} the cross entropy loss function tends to expand the variance between classes in the training process, which \deleted{will }cause\added{s} the model to ignore the hidden relationship between classes in the training process\replaced{.}{,} \added{The }shallow network will extract the overall semantic information of the target, and too concentrated\added{ a} weight value will lead to the loss of global semantic information.
	
	The training strategy of JGEtKD can more effectively retain the relationships between classes during training, thus capturing \deleted{the }hidden information \replaced{in }{among }the data. The \deleted{model we }trained \added{model }can obtain \replaced{a }{the }rough outline and hidden information of the target \replaced{in }{at }the shallow layer (relatively discrete distribution of weight values in the shallow layer), and then accurately capture key information in the deep network (relatively concentrated weight values in the deep network). This global first, then local observation is the same way humans perceive the world.
	\subsection{Ablation studies}
	
	\subsubsection{Knowledge distillation loss function}
	\begin{table}[h]
		\centering
		\caption{Performance of different knowledge distillation loss function in classification on ScanObjectNN.}\label{tab 3}
		\resizebox{0.6\linewidth}{!}{%
			\begin{tabular}{llccc}
				\hline
				\multirow{2}{*}{Teacher}      & \multirow{2}{*}{Student}         & \multirow{2}{*}{KD} & \multicolumn{2}{c}{ScanObjectNN} \\
				&                                  &                     & OA              & mAcc           \\ \hline
				\multirow{9}{*}{RepSurf-U\ddag2X} & \multirow{3}{*}{pointnet++\dag} & KD \cite{KD2015}                 & 80.57           & 79.14          \\
				&                                  & DKD \cite{DKD}                & 84.94           & 82.54          \\
				&                                  & \textbf{JGEtKD(ours)}        & 85.11           & 83.80          \\ \cline{2-5}
				& \multirow{3}{*}{RepSurf-U\ddag}  & KD \cite{KD2015}                 & 86.50           & 85.84          \\
				&                                  & DKD \cite{DKD}                & 88.31           & 87.54          \\
				&                                  & \textbf{JGEtKD(ours)}        & 88.62           & 87.85          \\ \cline{2-5}
				& \multirow{3}{*}{RepSurf-U\ddag2X}& KD \cite{KD2015}                 & 85.84           & 84.48          \\
				&                                  & DKD \cite{DKD}                 & 89.49           & 88.66          \\
				&                                  & \textbf{JGEtKD(ours)}        & \textbf{90.01}  & \textbf{88.81} \\ \hline
			\end{tabular}%
		}
	\end{table}
	To evaluate the performance of the different knowledge distillation loss \replaced{functions described }{function stated }in Section 2.1, we adopt the teacher knowledge distillation (tKD) strategy. We compared the performance of different knowledge distilled loss functions, and the results are \replaced{presented }{shown }in Table \hypersetup{linkcolor=cyan}\ref{tab 3} \replaced{, where}{. In the table,} KD is the standard distilled loss function \cite{KD2015}. Compared \replaced{to }{with }the advanced distillation strategy\added{ of} DKD \cite{DKD}, our approach demonstrate\replaced{d}{s} a difference of no less than 0.17\% in terms of OA.

	\subsubsection{Knowledge distillation framework}
	\begin{table}[h]
		\centering
		\caption{Performance of knowledge distillation framework in classification on ScanObjectNN.}\label{tab 4}
		\resizebox{0.6\linewidth}{!}{%
			\begin{tabular}{lcccc}
				\hline
				\multirow{2}{*}{Method}       & \multirow{2}{*}{Loss Function} & \multirow{2}{*}{Framework} & \multicolumn{2}{l}{ScanObjectNN} \\ \cline{4-5} 
				&                       &                            & OA              & mAcc           \\ \hline
				\multirow{3}{*}{RepSurf-U$\ddag$2X} & \textbf{JGEKD}                  & -                          & 86.33           & 85.90          \\ \cline{2-5} 
				& \textbf{JGEKD}                  & sKD                        & 89.24           & 88.34          \\ \cline{2-5} 
				& \textbf{JGEKD}                  & tKD                        & \textbf{90.01}  & \textbf{88.81} \\ \hline
			\end{tabular}
		}
	\end{table}
	To validate the effectiveness of the proposed framework, we compared the training performance\added{s} of RepSurf-U\ddag2X on the ScanObjectNN dataset using different frameworks. In Table \hypersetup{linkcolor=cyan}\ref{tab 4}, the model trained using our proposed framework (sKD\replaced{ and }{\&}tKD) \replaced{exhibited }{has }better performance while maintaining the loss function unchanged.

	\subsection{Robustness check}
	\begin{table*}[!h]
		\centering
		\caption{Average OA(\%) under 5 corruption levels of different model architectures on ModelNet40-C with different training strategies.}\label{tab 5}
		\resizebox{\linewidth}{!}{%
    	\begin{tabular}{llrrrrrrrrrrrrrrrr}             
		\hline
		\multicolumn{1}{l}{}                    & \multicolumn{1}{l}{} & \textbf{Rotation} & \textbf{Shear}    & \textbf{FFD}      & \textbf{RBF}      & \textbf{Lnv.RBF}  & \textbf{Uniform}  & \textbf{Ups.}     & \textbf{Gaussian} & \textbf{Impulse}  & \textbf{Bg.}      & \textbf{Occlusion} & \textbf{LiDAR}    & \textbf{Inc.}     & \textbf{Dec.}     & \textbf{Cutout}   & \textbf{m\_CE}    \\ \hline
		\multirow{6}{*}{\textbf{PointNet++}}     & \textbf{ST}          & 1.0{\color[HTML]{375623}(2)}            & 1.0{\color[HTML]{375623}(4)}            & 1.0{\color[HTML]{375623}(4)}            & 1.0{\color[HTML]{375623}(5)}            & 1.0{\color[HTML]{375623}(5)}            & 1.0{\color[HTML]{375623}(5)}            & 1.0{\color[HTML]{375623}(6)}            & 1.0{\color[HTML]{375623}(5)}            & 1.0{\color[HTML]{375623}(6)}            & 1.0{\color[HTML]{375623}(4)}            & 1.0{\color[HTML]{375623}(6)}             & 1.0{\color[HTML]{375623}(6)}            & 1.0{\color[HTML]{375623}(6)}            & 1.0{\color[HTML]{375623}(6)}            & 1.0{\color[HTML]{375623}(6)}            & 1.0{\color[HTML]{375623}(5)}            \\
		& \textbf{cutmix\_k}   & 1.04{\color[HTML]{375623}(3)}           & 0.894{\color[HTML]{375623}(2)}          & 0.913{\color[HTML]{375623}(2)}          & 0.857{\color[HTML]{375623}(2)}          & 0.902{\color[HTML]{375623}(2)}          & 0.982{\color[HTML]{375623}(4)}          & 0.906{\color[HTML]{375623}(2)}          & 0.977{\color[HTML]{375623}(4)}          & 0.46{\color[HTML]{375623}(4)}           & 0.799{\color[HTML]{375623}(3)}          & 0.876{\color[HTML]{375623}(3)}           & 0.796{\color[HTML]{375623}(3)}          & 0.406{\color[HTML]{375623}(4)}          & 0.511{\color[HTML]{375623}(2)}          & \textbf{0.569{\color[HTML]{375623}(1)}} & 0.793{\color[HTML]{375623}(2)}          \\
		& \textbf{cutmix\_r}   & 1.085{\color[HTML]{375623}(5)}          & 1.025{\color[HTML]{375623}(5)}          & 1.042{\color[HTML]{375623}(5)}          & 0.879{\color[HTML]{375623}(4)}          & 0.936{\color[HTML]{375623}(4)}          & 0.838{\color[HTML]{375623}(2)}          & 0.962{\color[HTML]{375623}(5)}          & 0.744{\color[HTML]{375623}(2)}          & 0.358{\color[HTML]{375623}(2)}          & 1.14{\color[HTML]{375623}(5)}           & 0.953{\color[HTML]{375623}(4)}           & 0.788{\color[HTML]{375623}(2)}          & 0.391{\color[HTML]{375623}(2)}          & 0.512{\color[HTML]{375623}(3)}          & 0.6{\color[HTML]{375623}(2)}            & 0.817{\color[HTML]{375623}(3)}          \\
		& \textbf{mixup}       & 1.068{\color[HTML]{375623}(4)}          & 0.923{\color[HTML]{375623}(3)}          & 0.955{\color[HTML]{375623}(3)}          & 0.872{\color[HTML]{375623}(3)}          & 0.905{\color[HTML]{375623}(3)}          & 0.912{\color[HTML]{375623}(3)}          & 0.917{\color[HTML]{375623}(3)}          & 0.868{\color[HTML]{375623}(3)}          & 0.55{\color[HTML]{375623}(5)}           & 1.751{\color[HTML]{375623}(6)}          & 0.964{\color[HTML]{375623}(5)}           & 0.996{\color[HTML]{375623}(5)}          & 0.562{\color[HTML]{375623}(5)}          & 0.747{\color[HTML]{375623}(5)}          & 0.751{\color[HTML]{375623}(5)}          & 0.916{\color[HTML]{375623}(4)}          \\
		& \textbf{rsmix}       & 1.819{\color[HTML]{375623}(6)}          & 1.305{\color[HTML]{375623}(6)}          & 1.45{\color[HTML]{375623}(6)}           & 1.562{\color[HTML]{375623}(6)}          & 1.604{\color[HTML]{375623}(6)}          & 1.897{\color[HTML]{375623}(6)}          & 0.919{\color[HTML]{375623}(4)}          & 2.098{\color[HTML]{375623}(6)}          & 0.433{\color[HTML]{375623}(3)}          & 0.725{\color[HTML]{375623}(2)}          & 0.82{\color[HTML]{375623}(2)}            & 0.799{\color[HTML]{375623}(4)}          & 0.404{\color[HTML]{375623}(3)}          & 0.524{\color[HTML]{375623}(4)}          & 0.612{\color[HTML]{375623}(4)}          & 1.131{\color[HTML]{375623}(6)}          \\
		& \textbf{DesenAT-sD}  & \textbf{0.755{\color[HTML]{375623}(1)}} & \textbf{0.555{\color[HTML]{375623}(1)}} & \textbf{0.808{\color[HTML]{375623}(1)}} & \textbf{0.733{\color[HTML]{375623}(1)}} & \textbf{0.765{\color[HTML]{375623}(1)}} & \textbf{0.677{\color[HTML]{375623}(1)}} & \textbf{0.866{\color[HTML]{375623}(1)}} & \textbf{0.586{\color[HTML]{375623}(1)}} & \textbf{0.331{\color[HTML]{375623}(1)}} & \textbf{0.52{\color[HTML]{375623}(1)}}  & \textbf{0.808{\color[HTML]{375623}(1)}}  & \textbf{0.673{\color[HTML]{375623}(1)}} & \textbf{0.38{\color[HTML]{375623}(1)}}  & \textbf{0.494{\color[HTML]{375623}(1)}} & 0.602{\color[HTML]{375623}(3)}          & \textbf{0.637{\color[HTML]{375623}(1)}} \\ \hline
		\multirow{6}{*}{\textbf{PointNetMeta-s}} & \textbf{ST}          & 0.961{\color[HTML]{375623}(2)}          & 0.922{\color[HTML]{375623}(3)}          & 0.961{\color[HTML]{375623}(4)}          & 0.948{\color[HTML]{375623}(4)}          & 0.999{\color[HTML]{375623}(4)}          & 0.952{\color[HTML]{375623}(3)}          & 0.955{\color[HTML]{375623}(6)}          & 0.922{\color[HTML]{375623}(3)}          & 1.228{\color[HTML]{375623}(6)}          & 0.881{\color[HTML]{375623}(5)}          & 0.867{\color[HTML]{375623}(3)}           & 0.934{\color[HTML]{375623}(6)}          & 0.426{\color[HTML]{375623}(6)}          & 0.902{\color[HTML]{375623}(6)}          & 0.923{\color[HTML]{375623}(6)}          & 0.919{\color[HTML]{375623}(5)}          \\
		& \textbf{cutmix\_k}   & 1.266{\color[HTML]{375623}(5)}          & 0.998{\color[HTML]{375623}(5)}          & 1.019{\color[HTML]{375623}(5)}          & 1.018{\color[HTML]{375623}(5)}          & 1.058{\color[HTML]{375623}(5)}          & 1.598{\color[HTML]{375623}(5)}          & 0.811{\color[HTML]{375623}(2)}          & 1.7{\color[HTML]{375623}(5)}            & 0.436{\color[HTML]{375623}(4)}          & 0.664{\color[HTML]{375623}(2)}          & \textbf{0.778{\color[HTML]{375623}(1)}}  & 0.738{\color[HTML]{375623}(3)}          & 0.315{\color[HTML]{375623}(3)}          & 0.498{\color[HTML]{375623}(3)}          & \textbf{0.539{\color[HTML]{375623}(1)}} & 0.896{\color[HTML]{375623}(4)}          \\
		& \textbf{cutmix\_r}   & 1.092{\color[HTML]{375623}(4)}          & 0.942{\color[HTML]{375623}(4)}          & 0.956{\color[HTML]{375623}(3)}          & 0.898{\color[HTML]{375623}(3)}          & 0.912{\color[HTML]{375623}(3)}          & 0.997{\color[HTML]{375623}(4)}          & \textbf{0.799{\color[HTML]{375623}(1)}} & 0.931{\color[HTML]{375623}(4)}          & \textbf{0.317{\color[HTML]{375623}(1)}} & 0.747{\color[HTML]{375623}(3)}          & 0.913{\color[HTML]{375623}(6)}           & \textbf{0.725{\color[HTML]{375623}(1)}} & \textbf{0.301{\color[HTML]{375623}(1)}} & \textbf{0.476{\color[HTML]{375623}(1)}} & 0.553{\color[HTML]{375623}(3)}          & 0.771{\color[HTML]{375623}(2)}          \\
		& \textbf{mixup}       & 0.999{\color[HTML]{375623}(3)}          & 0.829{\color[HTML]{375623}(2)}          & 0.897{\color[HTML]{375623}(2)}          & 0.848{\color[HTML]{375623}(2)}          & 0.885{\color[HTML]{375623}(2)}          & 0.92{\color[HTML]{375623}(2)}           & 0.892{\color[HTML]{375623}(5)}          & 0.84{\color[HTML]{375623}(2)}           & 0.445{\color[HTML]{375623}(5)}          & 1.377{\color[HTML]{375623}(6)}          & 0.901{\color[HTML]{375623}(4)}           & 0.899{\color[HTML]{375623}(5)}          & 0.38{\color[HTML]{375623}(5)}           & 0.662{\color[HTML]{375623}(5)}          & 0.676{\color[HTML]{375623}(5)}          & 0.83{\color[HTML]{375623}(3)}           \\
		& \textbf{rsmix}       & 1.606{\color[HTML]{375623}(6)}          & 1.071{\color[HTML]{375623}(6)}          & 1.238{\color[HTML]{375623}(6)}          & 1.347{\color[HTML]{375623}(6)}          & 1.376{\color[HTML]{375623}(6)}          & 2.034{\color[HTML]{375623}(6)}          & 0.82{\color[HTML]{375623}(3)}           & 2.148{\color[HTML]{375623}(6)}          & 0.342{\color[HTML]{375623}(3)}          & 0.791{\color[HTML]{375623}(4)}          & 0.903{\color[HTML]{375623}(5)}           & 0.736{\color[HTML]{375623}(2)}          & 0.313{\color[HTML]{375623}(2)}          & 0.477{\color[HTML]{375623}(2)}          & 0.539{\color[HTML]{375623}(2)}          & 1.049{\color[HTML]{375623}(6)}          \\
		& \textbf{DesenAT-sD}  & \textbf{0.939{\color[HTML]{375623}(1)}} & \textbf{0.556{\color[HTML]{375623}(1)}} & \textbf{0.896{\color[HTML]{375623}(1)}} & \textbf{0.813{\color[HTML]{375623}(1)}} & \textbf{0.856{\color[HTML]{375623}(1)}} & \textbf{0.671{\color[HTML]{375623}(1)}} & 0.87{\color[HTML]{375623}(4)}           & \textbf{0.565{\color[HTML]{375623}(1)}} & 0.322{\color[HTML]{375623}(2)}          & \textbf{0.457{\color[HTML]{375623}(1)}} & 0.807{\color[HTML]{375623}(2)}           & 0.8{\color[HTML]{375623}(4)}            & 0.35{\color[HTML]{375623}(4)}           & 0.558{\color[HTML]{375623}(4)}          & 0.602{\color[HTML]{375623}(4)}          & \textbf{0.671{\color[HTML]{375623}(1)}} \\ \hline
		\multirow{6}{*}{\textbf{APES\_global}}   & \textbf{ST}          & 0.784{\color[HTML]{375623}(3)}          & 0.673{\color[HTML]{375623}(2)}          & 0.792{\color[HTML]{375623}(3)}          & 0.878{\color[HTML]{375623}(5)}          & 0.905{\color[HTML]{375623}(5)}          & 0.892{\color[HTML]{375623}(4)}          & 0.852{\color[HTML]{375623}(3)}          & 0.873{\color[HTML]{375623}(5)}          & 0.711{\color[HTML]{375623}(6)}          & 0.444{\color[HTML]{375623}(3)}          & 0.973{\color[HTML]{375623}(3)}           & 0.98{\color[HTML]{375623}(6)}           & 0.389{\color[HTML]{375623}(4)}          & 0.709{\color[HTML]{375623}(6)}          & 0.828{\color[HTML]{375623}(6)}          & 0.779{\color[HTML]{375623}(4)}          \\
		& \textbf{cutmix\_k}   & 0.913{\color[HTML]{375623}(6)}          & 0.86{\color[HTML]{375623}(6)}           & 0.983{\color[HTML]{375623}(6)}          & 0.92{\color[HTML]{375623}(6)}           & 0.964{\color[HTML]{375623}(6)}          & 1.004{\color[HTML]{375623}(6)}          & 1.0{\color[HTML]{375623}(6)}            & 0.946{\color[HTML]{375623}(6)}          & 0.545{\color[HTML]{375623}(4)}          & 0.684{\color[HTML]{375623}(5)}          & 1.054{\color[HTML]{375623}(6)}           & 0.934{\color[HTML]{375623}(5)}          & 0.449{\color[HTML]{375623}(6)}          & 0.703{\color[HTML]{375623}(5)}          & 0.785{\color[HTML]{375623}(3)}          & 0.85{\color[HTML]{375623}(6)}           \\
		& \textbf{cutmix\_r}   & 0.865{\color[HTML]{375623}(5)}          & 0.837{\color[HTML]{375623}(5)}          & 0.93{\color[HTML]{375623}(5)}           & 0.835{\color[HTML]{375623}(3)}          & 0.873{\color[HTML]{375623}(3)}          & 0.809{\color[HTML]{375623}(2)}          & 0.894{\color[HTML]{375623}(4)}          & 0.752{\color[HTML]{375623}(2)}          & 0.436{\color[HTML]{375623}(2)}          & 0.563{\color[HTML]{375623}(4)}          & 0.979{\color[HTML]{375623}(4)}           & 0.887{\color[HTML]{375623}(4)}          & 0.361{\color[HTML]{375623}(3)}          & 0.579{\color[HTML]{375623}(3)}          & 0.827{\color[HTML]{375623}(5)}          & 0.762{\color[HTML]{375623}(3)}          \\
		& \textbf{mixup}       & 0.755{\color[HTML]{375623}(2)}          & 0.764{\color[HTML]{375623}(4)}          & 0.878{\color[HTML]{375623}(4)}          & 0.845{\color[HTML]{375623}(4)}          & 0.884{\color[HTML]{375623}(4)}          & 0.9{\color[HTML]{375623}(5)}            & 0.983{\color[HTML]{375623}(5)}          & 0.832{\color[HTML]{375623}(3)}          & 0.605{\color[HTML]{375623}(5)}          & 0.9{\color[HTML]{375623}(6)}            & 1.031{\color[HTML]{375623}(5)}           & 0.878{\color[HTML]{375623}(3)}          & 0.389{\color[HTML]{375623}(5)}          & 0.691{\color[HTML]{375623}(4)}          & 0.807{\color[HTML]{375623}(4)}          & 0.809{\color[HTML]{375623}(5)}          \\
		& \textbf{rsmix}       & 0.823{\color[HTML]{375623}(4)}          & 0.684{\color[HTML]{375623}(3)}          & 0.758{\color[HTML]{375623}(2)}          & 0.762{\color[HTML]{375623}(2)}          & 0.811{\color[HTML]{375623}(2)}          & 0.882{\color[HTML]{375623}(3)}          & 0.837{\color[HTML]{375623}(2)}          & 0.85{\color[HTML]{375623}(4)}           & 0.506{\color[HTML]{375623}(3)}          & 0.441{\color[HTML]{375623}(2)}          & 0.92{\color[HTML]{375623}(2)}            & \textbf{0.842{\color[HTML]{375623}(1)}} & \textbf{0.303{\color[HTML]{375623}(1)}} & \textbf{0.462{\color[HTML]{375623}(1)}} & \textbf{0.53{\color[HTML]{375623}(1)}}  & 0.694{\color[HTML]{375623}(2)}          \\
		& \textbf{DesenAT-sD}  & \textbf{0.619{\color[HTML]{375623}(1)}} & \textbf{0.497{\color[HTML]{375623}(1)}} & \textbf{0.677{\color[HTML]{375623}(1)}} & \textbf{0.689{\color[HTML]{375623}(1)}} & \textbf{0.723{\color[HTML]{375623}(1)}} & \textbf{0.603{\color[HTML]{375623}(1)}} & \textbf{0.778{\color[HTML]{375623}(1)}} & \textbf{0.529{\color[HTML]{375623}(1)}} & \textbf{0.322{\color[HTML]{375623}(1)}} & \textbf{0.369{\color[HTML]{375623}(1)}} & \textbf{0.903{\color[HTML]{375623}(1)}}  & 0.856{\color[HTML]{375623}(2)}          & 0.314{\color[HTML]{375623}(2)}          & 0.539{\color[HTML]{375623}(2)}          & 0.649{\color[HTML]{375623}(2)}          & \textbf{0.604{\color[HTML]{375623}(1)}} \\ \hline
		\multirow{6}{*}{\textbf{APES\_local}}    & \textbf{ST}          & 0.757{\color[HTML]{375623}(3)}          & 0.708{\color[HTML]{375623}(4)}          & 0.78{\color[HTML]{375623}(4)}           & 0.847{\color[HTML]{375623}(4)}          & 0.873{\color[HTML]{375623}(4)}          & 0.87{\color[HTML]{375623}(4)}           & 0.879{\color[HTML]{375623}(4)}          & 0.831{\color[HTML]{375623}(4)}          & 0.631{\color[HTML]{375623}(4)}          & 0.42{\color[HTML]{375623}(2)}           & 0.948{\color[HTML]{375623}(2)}           & 0.921{\color[HTML]{375623}(6)}          & 0.383{\color[HTML]{375623}(4)}          & 0.7{\color[HTML]{375623}(4)}            & 0.799{\color[HTML]{375623}(4)}          & 0.756{\color[HTML]{375623}(4)}          \\
		& \textbf{cutmix\_k}   & 0.984{\color[HTML]{375623}(6)}          & 1.011{\color[HTML]{375623}(5)}          & 1.161{\color[HTML]{375623}(6)}          & 1.082{\color[HTML]{375623}(6)}          & 1.115{\color[HTML]{375623}(6)}          & 1.217{\color[HTML]{375623}(6)}          & 1.246{\color[HTML]{375623}(6)}          & 1.128{\color[HTML]{375623}(6)}          & 0.703{\color[HTML]{375623}(6)}          & 1.066{\color[HTML]{375623}(6)}          & 0.997{\color[HTML]{375623}(5)}           & 0.869{\color[HTML]{375623}(4)}          & 0.492{\color[HTML]{375623}(6)}          & 0.807{\color[HTML]{375623}(6)}          & 0.876{\color[HTML]{375623}(5)}          & 0.984{\color[HTML]{375623}(6)}          \\
		& \textbf{cutmix\_r}   & 0.96{\color[HTML]{375623}(5)}           & 1.048{\color[HTML]{375623}(6)}          & 1.07{\color[HTML]{375623}(5)}           & 1.005{\color[HTML]{375623}(5)}          & 1.069{\color[HTML]{375623}(5)}          & 1.044{\color[HTML]{375623}(5)}          & 1.215{\color[HTML]{375623}(5)}          & 0.974{\color[HTML]{375623}(5)}          & 0.642{\color[HTML]{375623}(5)}          & 0.982{\color[HTML]{375623}(5)}          & 1.062{\color[HTML]{375623}(6)}           & 0.88{\color[HTML]{375623}(5)}           & 0.443{\color[HTML]{375623}(5)}          & 0.732{\color[HTML]{375623}(5)}          & 0.876{\color[HTML]{375623}(6)}          & 0.933{\color[HTML]{375623}(5)}          \\
		& \textbf{mixup}       & \textbf{0.656{\color[HTML]{375623}(1)}} & 0.645{\color[HTML]{375623}(2)}          & 0.756{\color[HTML]{375623}(2)}          & 0.736{\color[HTML]{375623}(2)}          & 0.761{\color[HTML]{375623}(2)}          & 0.757{\color[HTML]{375623}(2)}          & 0.875{\color[HTML]{375623}(3)}          & 0.668{\color[HTML]{375623}(2)}          & 0.474{\color[HTML]{375623}(3)}          & 0.734{\color[HTML]{375623}(4)}          & 0.959{\color[HTML]{375623}(4)}           & 0.854{\color[HTML]{375623}(3)}          & 0.353{\color[HTML]{375623}(3)}          & 0.617{\color[HTML]{375623}(3)}          & 0.655{\color[HTML]{375623}(2)}          & 0.7{\color[HTML]{375623}(3)}            \\
		& \textbf{rsmix}       & 0.848{\color[HTML]{375623}(4)}          & 0.675{\color[HTML]{375623}(3)}          & 0.758{\color[HTML]{375623}(3)}          & 0.745{\color[HTML]{375623}(3)}          & 0.774{\color[HTML]{375623}(3)}          & 0.862{\color[HTML]{375623}(3)}          & 0.843{\color[HTML]{375623}(2)}          & 0.809{\color[HTML]{375623}(3)}          & 0.43{\color[HTML]{375623}(2)}           & 0.446{\color[HTML]{375623}(3)}          & 0.957{\color[HTML]{375623}(3)}           & \textbf{0.835{\color[HTML]{375623}(1)}} & \textbf{0.303{\color[HTML]{375623}(1)}} & \textbf{0.467{\color[HTML]{375623}(1)}} & \textbf{0.536{\color[HTML]{375623}(1)}} & 0.686{\color[HTML]{375623}(2)}          \\
		& \textbf{DesenAT-sD}  & 0.709{\color[HTML]{375623}(2)}          & \textbf{0.52{\color[HTML]{375623}(1)}}  & \textbf{0.718{\color[HTML]{375623}(1)}} & \textbf{0.696{\color[HTML]{375623}(1)}} & \textbf{0.716{\color[HTML]{375623}(1)}} & \textbf{0.65{\color[HTML]{375623}(1)}}  & \textbf{0.813{\color[HTML]{375623}(1)}} & \textbf{0.576{\color[HTML]{375623}(1)}} & \textbf{0.373{\color[HTML]{375623}(1)}} & \textbf{0.391{\color[HTML]{375623}(1)}} & \textbf{0.925{\color[HTML]{375623}(1)}}  & 0.853{\color[HTML]{375623}(2)}          & 0.339{\color[HTML]{375623}(2)}          & 0.573{\color[HTML]{375623}(2)}          & 0.674{\color[HTML]{375623}(3)}          & \textbf{0.635{\color[HTML]{375623}(1)}} \\ \hline
	\end{tabular}
		}
		\raggedright
		\footnotesize Bg.: Background, Ups.: Upsampling.
	\end{table*}
	\replaced{As shown in}{In} Table \hypersetup{linkcolor=cyan}\ref{tab 5}, our experiments on corruption robustness were conducted on all benchmarks of ModelNet40-C. \replaced{To }{In order to }evaluate the performance of our proposed strategies (JGEsKD and JGEtKD), in Section 4.5.1, we compared JGEsDK and JGEtKD with \added{the }ST and existing point cloud data augmentation strategies. \replaced{then, in }{In }Section 4.5.2, we evaluate the robustness of each model as the severity of corruption increases under different training strategies.
	\subsubsection{Robustness comparison experiment}
%	\begin{figure*}[ht]
%		\centering 
%		\includegraphics[width=0.7\linewidth]{radar_chart.eps}\par %555 radar_chart
%		\caption{Radar chart: Average OA(\%) under 5 corruption levels of different models under different training strategies in ModelNet40-C dataset.}
%		\label{radar_chart}
%	\end{figure*}
%	Fig. \hypersetup{linkcolor=cyan}\ref{radar_chart} \replaced{presents }{shows }a radar chart of the corruption robustness of different models under different training strategies. \replaced{The figure shows }{From the figure, it can be observed }that JGEsKD and JGEtKD are mostly located in the outer ring for most benchmarks, indicating that the overall robustness of our proposed strategy is superior to\added{ that of the} existing training strategies.

	\replaced{An }{We also observed an }interesting phenomenon\added{ is observed}. Taking RepSurf-U$\ddag$2x as an example, under the "Transformation Corruption" benchmark\deleted{s}, the OA obtained by our proposed strategy remains stable \replaced{at approximately }{around }91\% for all \replaced{five}{5} benchmarks, while the existing data augmentation strategy Mixup \cite{Pointmixup} can only achieve an OA of 80.74\% to 89.89\%. This may be \replaced{becaused }{due to the fact that }our strategy maintains data augmentation throughout the training process, which has certain resistance to data from perturbations. In the "Noise Corruption" benchmarks, our strategy outperforms other training strategies in "Uniform", "Gaussian", "Impulse" and "Upsampling" benchmarks. However, \added{the performance of }our strategy\deleted{'s performance} falls short \added{of that }of Cutmix\_k and Cutmix\_r \cite{Pointcutmix} on the "Background" (Bg.) benchmark. This may be because Cutmix\_k and Cutmix\_r employ sample superposition to make\added{ the} samples mutually backgrounds, which effectively simulates the corruption scenario and improves the robustness of the model. \replaced{By }{In }contrast, our strategy induces the model to learn different representations of the same sample, while ignoring \added{the }background, which leads to lower robustness.  Finally, in the "Density Corruption" benchmarks, although the OA is not high for "Occlusion" and "LIDAR", our proposed strategy still maintains\added{ a} relatively high performance compared \replaced{with }{to }other training strategies.
%	\begin{figure*}[!h]
%		\centering 
%		\includegraphics[width=0.75\linewidth]{line.eps}
%		\caption{Different levels of robust OA of different models with different data augmentation strategies on modelNet40-C under different robustness levels.} 
%		\label{line}
%	\end{figure*}
	
	\begin{figure*}[!h]
		\centering 
		\includegraphics[width=0.8\linewidth]{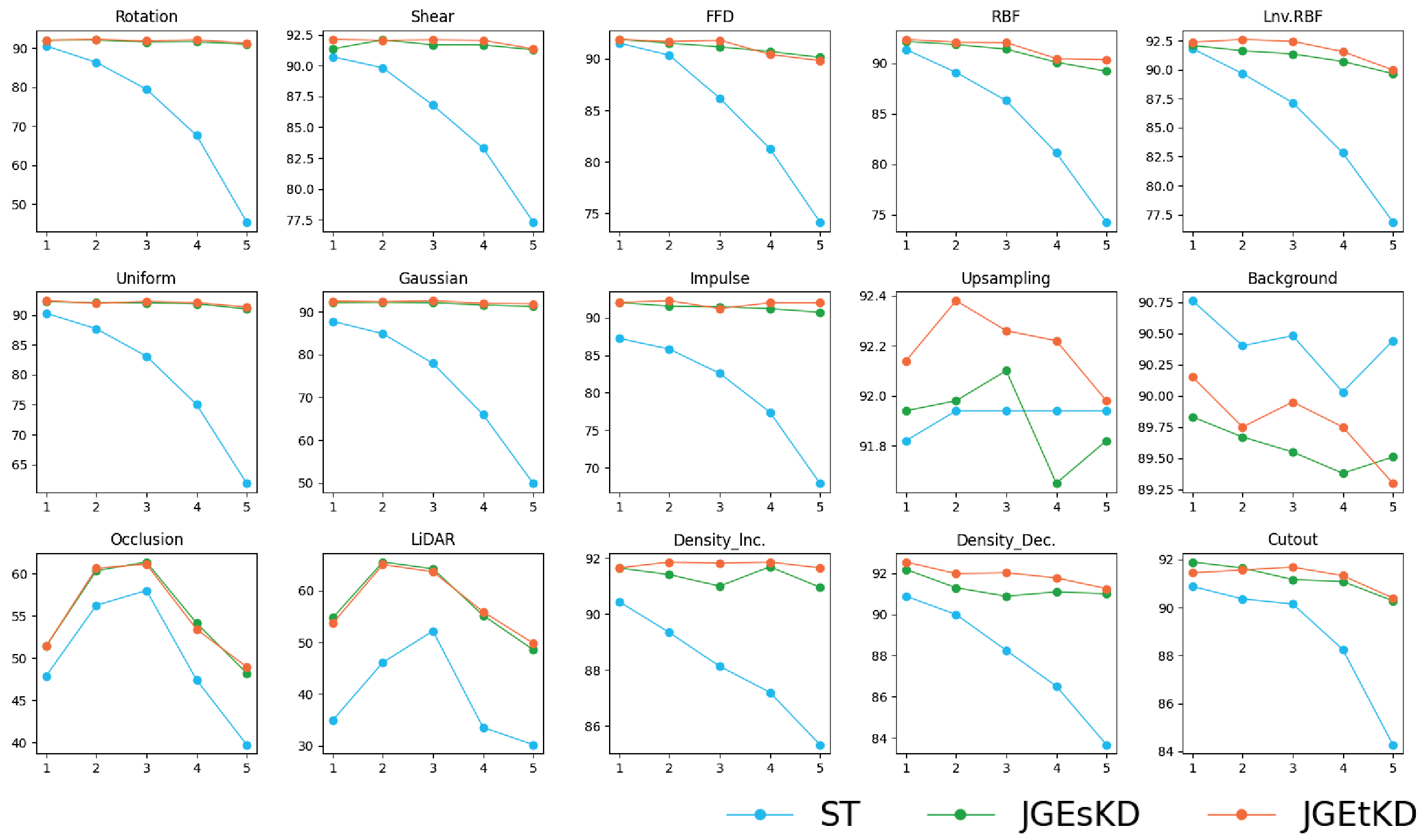}
		\caption{OA of PointNeXt-S with different data augmentation strategies on modelNet40-C under different robustness levels.} 
		\label{pxtline}
	\end{figure*}
	\subsubsection{Robustness level test}
	\replaced{As shown in }{In }Fig\added{s}. \deleted{Fig. }\ref{pxtline}, we conducted experiments to evaluate the performance of different models with different training strategies at various robustness levels. \replaced{As }{It can be observed that, as }the robustness level increases, the overall OA of the models exhibits a decreasing trend. Compared \replaced{with the }{to }ST and existing training strategies, our proposed strategies (JGEsKD\added{ and} JGEtKD) are generally more stable. Notably, our strategies show higher robustness than all the compared strategies in all benchmarks of "Transformation Corruptions" and some benchmarks "Occlusion", "Uniform", "Gaussian", "LiDAR" of "Noise Corruptions" and "Density Corruptions", reaching a leading level. However, we found that our strategies are not as effective as the current sample fusion data augmentation strategy on some benchmarks of changing point numbers. This may be \replaced{because }{due to the fact that }our training strategy did not \replaced{consider changes in the number of points}{take into account the changes in point numbers}.
	\section{Conclusion}
	This \replaced{study}{work} re-examines point cloud classification tasks from a non-IID perspective, positing that \deleted{there exist }hidden relationships between different class events. To address this, we propose the employ of a joint graph to construct these relationships and introduce a graph knowledge distillation strategy to facilitate information transmission across different representations of the same sample. Based on this strategy, we construct two frameworks, JGEsKD and JGEtKD, to meet the needs of different scenarios. Compared \replaced{with }{to }the baseline RepSurf-U\ddag2X, the OA metrics increase by 3.99\% on the ScanObject dataset. Finally, we utilize the JGEKD framework for anti-corruption training, which effectively enhances the \deleted{model's }robustness\added{ of the model} against corruption.

	Although the strategy in this \replaced{study }{paper }is simple and effective \replaced{in }{to }improve model performance without increasing the computational load, this method has two shortcomings: (1)it only considers the non-IID relationship between classes; in fact, \replaced{These}{the samples also These} implicit relationships should \added{also }exist. (2) \replaced{No }{During the training process, no }additional datasets are introduced\added{ during training}. In \replaced{future studies}{further research}, we \replaced{plan to consider this }{will consider the }problem from these two \replaced{perspectives}{aspects}. 

	\replaced{P}{In the future, p}oint cloud classification and robustness research will remain \deleted{a }challeng\replaced{e}{ing}, \replaced{bacause }{as }real-world point clouds are complex and diverse, and existing datasets may not adequately represent all scenarios. Therefore, it is crucial to implement transfer learning methods to enhance model recognition and robustness.
	\section*{Acknowledgment}
	This work was partially supported by the National Natural Science Foundation of China under Grants (No.51774219), and the key R\&D Program of Hubei Province (No.2020BAB098).
	
	Numerical calculation is supported by High-Performance Computing Center of Wuhan University of Science and Technology.
	
	\bibliographystyle{unsrt}
	\bibliography{ours.bib}
	
\end{document}